# Do Robots Need Body Language? Comparing Communication Modalities for Legible Motion Intent in Human-Shared Spaces

Jonathan Albert Cohen*
jonny@media.mit.edu
MIT Media Lab
Cambridge USA

Kye Shimizu*
kyeshmz@media.mit.edu
MIT Media Lab
Cambridge USA

Allen Song†
allen017@media.mit.edu
MIT Media Lab
Cambridge USA

Vishnu Bharath†
vishnu09bharath@gmail.com
New England Innovation Academy
Cambridge USA

Pattie Maes
pattie@media.mit.edu
MIT Media Lab
Cambridge USA

Kent Larson
kll@media.mit.edu
MIT Media Lab
Cambridge USA

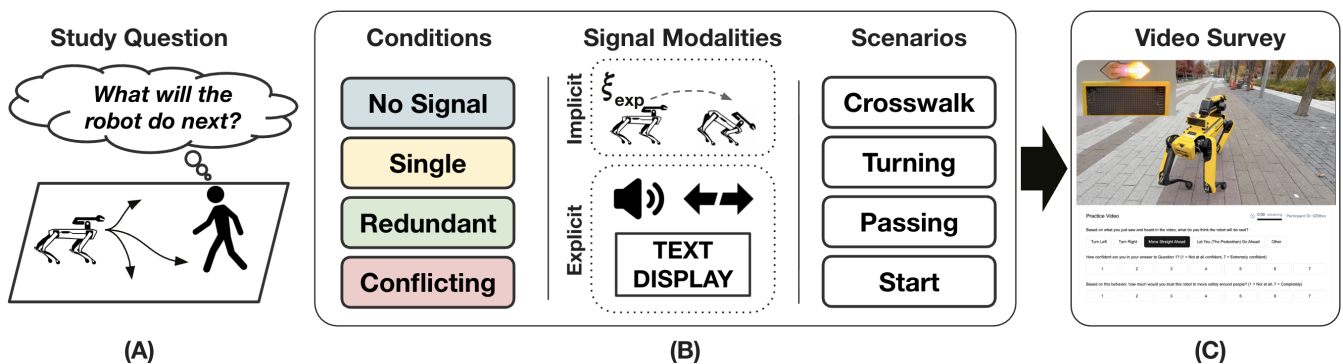

Figure 1: Study design for evaluating how different robot signaling strategies support human prediction of motion intent. (A) Participants inferred a quadruped robot's next action after viewing short videos. (B) Each video applied one of four signal conditions (no signal, single, redundant, conflicting) implemented through different signaling modalities, implicit expressive motion trajectories ($\xi_{\text{exp}}$) and explicit channels (lights, text, audio), across four navigation scenarios (crosswalk, turning, passing, starting movement). In total, 78 unique videos were produced. (C) In the online survey, 210 participants viewed ten randomized clips and reported the robot's predicted next action, as well as their confidence and trust ratings.

## Abstract

Robots in shared spaces often move in ways that are difficult for people to interpret, placing the burden on humans to adapt. High-DoF robots exhibit motion that people read as expressive, intentionally or not, making it important to understand how such cues are perceived. We present an online video study evaluating how different signaling modalities, expressive motion, lights, text, and audio, shape people's ability to understand a quadruped robot's upcoming navigation actions (Boston Dynamics Spot [4]). Across four common scenarios, we measure how each modality influences humans' (1) accuracy in predicting the robot's next navigation action, (2) confidence in that prediction, and (3) trust in the robot to act safely. The study tests how expressive motions compare to explicit channels, whether aligned multimodal cues enhance interpretability, and how conflicting cues affect user confidence and trust. We contribute initial evidence on the relative effectiveness of implicit versus explicit signaling strategies.

---

*Authors contributed equally to the paper
†Second authors

# 1 Introduction

Legged robots are being used in human-shared spaces, where people must quickly infer what the robot will do next in order to stay and feel safe. In these environments, robots should not place the burden on nearby pedestrians to avoid them; in human-centric spaces like sidewalks, it is the robot's responsibility to communicate its motion clearly and predictably. A robot that suddenly pivots or accelerates without warning can startle people or force avoidance [30], even if no actual collision occurs. Prior work on legible robot motion formalizes this challenge as an inference problem: motion should be planned not only to achieve a goal, but also to make that goal easy for humans to infer from partial observations of the trajectory [8]. Yet, in most everyday encounters, human observers have access to more than just the robot's path, they also see its posture, lights, or hear sound and speech [1].

As robots gain higher degrees of freedom, their movements often communicate information, whether intentionally designed or not. Expressive body motions (*body language*) arise from the robot's morphology and control, require no additional hardware, and remain visible from multiple viewpoints. Yet these cues can be ambiguous or misleading, especially when a robot pursues multiple goals simultaneously (e.g., navigating while manipulating an object unaligned with the robot heading) [7, 22]. Understanding how people interpret such implicit motion cues is therefore essential for designing legible, predictable, and trustworthy robot behavior in human-shared environments.

Prior HRI research has shown that animation-inspired motion can make robot actions more readable and improve subjective perceptions of confidence and intelligence [32], and that expressive motion can convey their limitations and failures [20] or span a broader design space of expressive behaviors [12, 31]. Yet there is still limited empirical understanding of how different communication *channels*—text, audio, lights, and body language—compare on the same platform and task, or how they interact when they agree or conflict.

In this paper, we present a study that systematically compares four channels for conveying a legged robot's next navigation action: (1) on-body text, (2) synthesized audio, (3) indicator lights, and (4) expressive body motions. Participants watch short clips of a quadruped robot navigating human-shared scenarios, then predict the robot's next action, rate their confidence, and report their trust. We recruited 210 participants through Prolific [27] (50.5% men, 48.0% women, 1.5% non-binary; mean age = 41.91, SD = 11.93; 64% White, 15% Asian, 11% Black; 68% college-educated). Our study addresses the research question: **How do implicit expressive body motion cues compare to explicit modalities (lights, text, audio), when used individually, redundantly, or in conflict, in supporting accurate prediction of a robot's next action and shaping user confidence and trust?**

We hypothesize that (H1) expressive body motion improves next-action prediction accuracy, confidence, and trust relative to no signaling; (H2) combining expressive motion with aligned explicit signals yields higher performance than any single channel; and (H3) conflicts between expressive motion and other channels degrade confidence and trust relative to aligned conditions.

Our contributions are: a combinatorial experimental design that compares text, audio, lights, and expressive body motion on a common prediction task, and an analysis of cross-channel agreement and conflict effects on prediction accuracy, confidence, and trust.

# 2 Related Work

**Legible and expressive robot motion**: Dragan et al. introduced a formal distinction between *predictable* and *legible* motion, where predictable trajectories are those that match an observer's expectations, while legible trajectories are those that make the robot's goal easy to infer from partial motion [8]. Subsequent work has explored legibility metrics, observer models, and legible planning in multi-robot and manipulation contexts [28], including learning observer models to produce legible trajectories [37], systematic comparisons of legibility frameworks [36], and extensions to multi-robot systems [3]. Beyond legibility, recent work has begun to explicitly treat motion as an expressive resource. Kwon et al. study how robots can communicate *incapability* through their motion when they fail to complete a task [20], while Sripathy et al. propose techniques for teaching robots to span a broader space of functional expressive motions [31]. In autonomous driving, Sadigh et al. show that planning that explicitly reasons about how human drivers respond to the car's motion can yield more effective and communicative interactions [30].

**Animation Principles and Robot Body Language**: Takayama et al. apply anticipation and reaction to robot behavior, showing that forethought and post-task reactions can improve readability, perceived competence, and approachability [32]. Hoffman and Ju argue that robots should be designed "with movement in mind," treating motion as a primary design material and emphasizing expressivity and character movement design [18]. These works suggest that nonverbal motion cues can shape how observers interpret what a robot is doing and what it will do next. Expressive motion has also been used to communicate internal states such as confusion or failure [20], and to shape the qualitative "style" of movement in service of affect or personality rather than strictly functional goals [31]. Recent design guidance for quadrupeds further emphasizes the value of anticipatory whole-body motions to externalize navigation intent and increase predictability [17]. For legged platforms, expressive body language can emerge from base orientation, gait modulation [6], and whole-body posture, but relatively few empirical studies have isolated these parameters as *signals* about upcoming navigation actions in human-shared spaces.

**Communicating Motion Intent Across Channels**: Pascher et al.'s scoping review synthesizes techniques for communicating robot motion intent, cataloguing methods that range from trajectory exaggeration and gaze cues to visual displays, lights, and AR overlays [25]. Research explores new channels: head-mounted displays for arm motion intent [29], projected paths on the ground, or light-based signaling systems that convey direction. In driver–vehicle interaction and autonomous vehicle research, expressive driver–vehicle interfaces and biomimetic vehicle-to-pedestrian communication protocols (e.g., AEVITA) explore how vehicles might signal yielding, intent to start moving, or anticipated paths through lights, surfaces, and motion patterns, often inspired by biological motion [23, 26, 38].

Closer to our domain, quadruped-specific HRI work has examined perceived safety and comfort. Hashimoto et al.'s Safe Spot study compares dominant versus submissive quadruped body postures and their effect on perceived safety [15], while Erlebach et al. investigate human comfort levels with quadruped robots at varying distances and approach angles in shared spaces [9]. These studies show that body posture and proxemic behavior strongly influence how people *feel* around legged robots [33], but they do not directly measure how these cues support prediction of concrete next actions (e.g., turning left versus going straight). At the planning level, social navigation frameworks treat human motion as part of the environment, modeling local interactions using social force models [16] or probabilistic crowd models for dense environments [35]. These methods often implicitly assume that the robot's future motion will be inferred from its current path [11, 13, 35].

**Positioning Our Work**: In summary, prior work has: (i) formalized legible motion and demonstrated that trajectories can be optimized to make goals easier to infer [8], (ii) shown that animation principles and expressive motion can improve readability and subjective impressions [18–20, 31, 32], (iii) catalogued a broad design space of motion-intent communication techniques [25], and (iv) begun to map human comfort and safety perceptions around quadruped robots [9, 15]. However, it is unclear how different communication channels compare when conveying the *same* navigation intent on a robot, or how cross-channel conflicts influence prediction, confidence, and trust. Our study addresses this gap by experimentally comparing expressive body motion, lights, text, and audio [5, 24].

## 3 Methods
### 3.1 Signal Design
We compared how four communication modalities, including a no-signaling baseline, convey a robot's next navigation action: **body language**, **lights**, **text**, and **audio** tested across the four scenarios (*crosswalk*, *turning*, *passing*, *starting*), while the robot's base trajectory remained consistent within a scenario. We implemented a custom HMI for multimodal signaling that was mounted visible and enlarged in each video. Expressive motion cues were informed by prior work on animation principles and HRI using whole-body and limb signals. Directional leaning and limb movements can support anticipatory legibility [2, 17, 21, 32], while deferential postures such as bowing or crouching align with findings on submissive quadruped behavior and perceived yielding [9, 15]. Explicit cues (text, audio, lights) were selected as widely studied intent-communication strategies for robots [10, 21, 25].

**Expressive Motion**: Whole-body postures and limb movements provided implicit expressive cues ($\xi_{\text{exp}}$). Directional leaning and limb gestures (*yaw/roll/arm left/right*) indicated turning; forward emphasis (*pitch_up, body_extend, arm_extend*) indicated movement ahead; and deferential postures (*bow, crouch, sit*) signaled yielding.

**Lights**: LED lights from the display conveyed robot yielding. Directional lights (*left/right*) indicated turning.

**Text and Audio**: Concise textual and spoken phrases ("TURNING LEFT/RIGHT", "PEDESTRIAN CROSSING", "ROBOT CROSSING", "PEDESTRIAN PASSING") explicitly stated the upcoming action using the same phrase.

**Cross-Channel Conditions**: Signals appeared as: (1) **single-channel** (e.g., *yaw_left*), (2) **redundant** when multiple modalities aligned (e.g., *yaw_left + light_left*), or (3) **conflicting** when modalities expressed opposing intents (e.g., *text_turning_right + yaw_left*). This permitted analysis of implicit-only communication, benefits of aligned multimodality, and the impact of misalignment.

### 3.2 Rating Task
During the evaluation phase, each participant viewed ten robot navigation videos (drawn from 78 total clips; M = 9.74s, SD = 1.28s). After each video, participants responded to three questions assessing: **(Q1): Prediction:** Based on what you just saw and heard, what do you think the robot will do next? Options are: Turn Left; Turn Right; Move Straight Ahead; Let the Pedestrian Go Ahead; Stop or Remain Stopped; Other. **(Q2): Confidence:** How confident are you in your answer to Q1? (1–7 Likert). **(Q3): Trust:** Based on this behavior, how much would you trust this robot to move safely around people? (1–7 Likert). This study was approved by the Institutional Review Board of MIT (IRB Protocol #2509001789).

### 3.3 Procedure
The study followed a four-phase procedure. Demographic information (age, gender, education) was collected automatically by the survey platform. **(1) Consent:** Participants reviewed the study description and provided informed consent. **(2) Instructions:** Participants were introduced to the robot and signaling modalities via a diagram and example question. **(3) Video Task:** Participants viewed ten randomized navigation videos and answered the three evaluation questions after each; one attention check was included. **(4) Post-Survey:** Participants completed final questions about their overall impressions of the robot and study.

## 4 Results
### 4.1 Data Analysis
We analyzed participants' prediction accuracy (Q1), confidence (Q2), and trust ratings (Q3) across signaling modalities and conditions. Each video belonged to one of nine signal categories: *none*, *body language (implicit)*, *light*, *audio*, *text*, *implicit + explicit (redundant)*, *explicit + explicit (redundant)*, *implicit vs. explicit (conflict)*, and *explicit vs. explicit (conflict)*. We recruited 210 participants who each viewed 10 video clips. Accounting for within-participant clustering, a sensitivity analysis indicated 80% power to detect effects of $f \geq 0.16$ at $\alpha = .05$. We computed participant-level mean accuracy, confidence and trust ratings, then compared each condition to the baseline using paired t-tests with Holm-corrected p-values.

**Accuracy Across Modalities**: Prediction accuracy varied widely across modalities (Fig. 2) and with all signal conditions significantly outperforming the no-signal baseline (p<.001). Baseline trials with no signal produced the lowest accuracy (≈14%), indicating that participants struggled to infer the robot's next action from context alone when no communicative cues were present. Implicit expressive body motion improved accuracy substantially (≈44%), demonstrating that whole-body expressive cues, even without optimization for legibility, provide meaningful information. Explicit modalities yielded the strongest performance: audio (≈82%) and text (≈88%) significantly outperformed other modalities. Lights

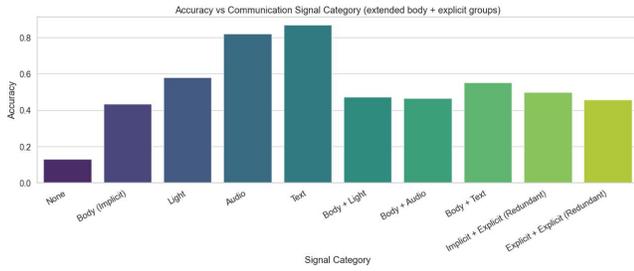

Figure 2: Accuracy vs. Communication Signal Category

produced moderate accuracy (≈58%). Redundancy did not increase accuracy for any explicit channels.

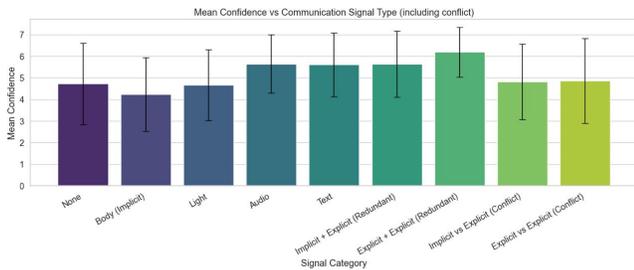

Figure 3: Mean Confidence vs. Communication Signal Type

**Confidence Ratings**: Mean confidence ratings (Fig. 3) followed a similar trend, with Body, Audio, Text, and Implicit+Explicit (redundant) conditions differing significantly from the no-signal baseline. Participants in the no-signal condition expressed moderate confidence ($M \approx 4.7$). Body Motion cues slightly reduced confidence relative to baseline ($M \approx 4.2$), suggesting that while the cues were informative enough to improve accuracy, they were not consistently perceived as reliable. Explicit modalities generated higher confidence: audio and text both produced means above 5.5, and redundant explicit signals elicited the highest confidence ($M \approx 6.1$). Conflict conditions lowered confidence, with implicit–explicit conflicts showing a sharper drop than explicit–explicit conflicts.

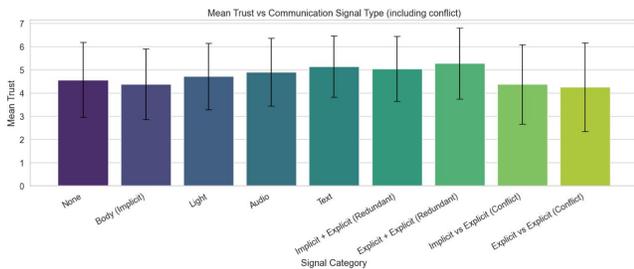

Figure 4: Mean Trust vs. Communication Signal Type

**Trust Ratings**: Trust ratings (Fig. 4) showed the same broad structure, but differences between conditions were not statistically significant. Participants trusted the robot least when it provided no signal ($M \approx 4.5$). Implicit body motion did not improve the trust rating or reach the levels of explicit cues. Text and audio achieved the highest trust scores (above 5), while redundant signals were similar. Additionally, Conflict conditions reduced trust slightly.

## 5 Discussion and Conclusion

**Body Motion Outperforms the Baseline**: Expressive body motion, even in a non-optimized form, substantially outperformed the no-signal baseline in accuracy (≈44% vs ≈14%) when the robot expressed directional leaning, gait cues, or deferential postures. This supports (H1), although confidence and trust did not increase above no signaling. Many robot designs lack explicit HMIs altogether; our data suggest that intentionally shaping inherent whole-body motion is a low-cost, high-impact pathway for improving legibility.

**Explicit Modalities Are Strong**: Text and audio produced the highest accuracy, confidence, and trust scores. This is expected: language is unambiguous and directly communicates intent. However, these modalities face practical challenges, including language diversity, ambient noise, vantage-point limitations, and visual clutter. Lights, a more universal signaling convention, ranked below text/audio but achieved performance comparable to or slightly above body language.

**Redundant Signals Are Limited**: Aligned multimodal signals produced slight improvements in confidence and trust compared to single explicit channels, partially supporting the hypothesis (H2). This ceiling effect may reflect added cognitive load from processing multiple simultaneous channels, limiting the informational benefit of redundancy.

**Conflicts Have Less Impact**: Conflict conditions did not reduce confidence relative to the no-signal baseline and produced only a slight decrease in trust. These results refine our hypothesis (H3): while we expected conflicts to meaningfully erode confidence and trust, their overall impact was smaller than anticipated.

**Limitations**: Key limitations include online video stimuli (which may overestimate explicit signal effectiveness relative to in-person interactions), a single robot platform, non-optimized motion trajectories, and unmodeled cultural differences. Future work should validate findings in real settings across morphologies and cultures, incorporating cognitive load measures [14].

**Conclusion:** Overall, expressive motion improves prediction versus no signaling, yet still falls short of the clarity offered by explicit channels such as text and audio. Even so, expressive motion remains a valuable modality: it is always visible, does not rely on language, and is compatible with many robots. These results suggest that carefully designed expressive motion has potential for enhancing legibility in human-shared environments and warrants further exploration as a core layer of robot communication.

## Acknowledgments

We thank the Media Lab for Sandbox support, Naroa Coretti Sánchez for experiment advice, Lauren Safier for filming assistance, and Pioneer Material Precision Tech Co., Ltd. for the Boston Dynamics Spot robot.